\documentclass[journal,onecolumn]{IEEEtran}

\usepackage{xcolor,soul,framed} %,caption
\usepackage{amsfonts}
\usepackage[pdftex]{graphicx}
\graphicspath{{../pdf/}{../jpeg/}}
\DeclareGraphicsExtensions{.pdf,.jpeg,.png}
\usepackage{amsmath,bm}
\usepackage{amsfonts}
\usepackage{tikz,float}
\usepackage{multirow}
\usepackage{adjustbox}
\usepackage{array}
\usepackage{mdwmath}
\usepackage{graphicx}
\usepackage{tabularx}
\usepackage{mdwtab}
\usepackage{eqparbox}
\usepackage{url}
\usepackage{caption}
\begin{document}
\bstctlcite{IEEEexample:BSTcontrol}
    \title{Efficient Diffusion Models for Vision: A Survey}
  \author{Anwaar Ulhaq,~\IEEEmembership{Member,~IEEE,} and 
      Naveed Akhtar,~\IEEEmembership{Member,~IEEE}
     
     % <-this % stops a space

 %= \thanks{Manuscript received July 10, 2012. \hl{This paper is an %=expanded paper from the IEEE MTT-S Int. Microwave Symposium held on %=June 17-22, 2012 in Montreal, Canada.} This work was funded in part by the Office of Naval Research under the Defense Advanced Research %=Projects Agency (DARPA) Microscale Power Conversion (MPC) Program under Grant N00014-11-1-0931, and in part by the Advanced Research Projects Agency-Energy (ARPA-E), U.S. Department of Energy, under Award Number DE-AR0000216.}
 %= \thanks{M. Roberg is with TriQuint Semiconductor, 500 West Renner Road Richardson, TX 75080 USA (e-mail: michael.roberg@tqs.com).}% %=<-this % stops a space
  %\thanks{* Both authors have equally contributed to the article during all stages of manuscript writing, preparation, and literature analysis.}%
%=  \thanks{I. Ramos and Z. Popovic are with the Department of Electrical, Computer and Energy Engineering, University of Colorado, Boulder, CO, 80309-0425 USA (e-mail: ignacio.ramos@colorado.edu; %=zoya.popovic@colorado.edu).}% <-this % stops a space
%=  \thanks{E. Falkenstein is with Qualcomm Inc., 6150 Lookout Road
%=Boulder, CO 80301 USA (e-mail: erez.falkenstein@gmail.com).}
}

\markboth{IEEE TRANSACTIONS ON XXXXXXXXXX
}{Roberg \MakeLowercase{\textit{et al.}}: High-Efficiency Diode and Transistor Rectifiers}

% === ABSTRACT ====================================================================
% =================================================================================

\maketitle

\begin{abstract}
%\boldmath
Diffusion Models (DMs) have demonstrated state-of-the-art performance in content generation without requiring adversarial training. These models are trained using a two-step process. First, a forward - \emph{diffusion} - process gradually adds noise to a datum (usually an image). Then, a backward - \emph{reverse diffusion} - process gradually removes the noise to turn it into a sample of the target distribution being modelled.. DMs are inspired by non-equilibrium thermodynamics and have inherent high computational complexity. Due to the frequent function evaluations and gradient calculations in high-dimensional spaces, these models incur considerable computational overhead during both  training and inference stages. This can not only preclude the democratization of diffusion-based modelling, but also hinder the adaption of diffusion models in real-life applications. Not to mention, the efficiency of computational models is fast becoming a significant concern due to excessive energy consumption and environmental scares. These factors have led to multiple contributions in the literature that focus on devising computationally efficient DMs.
In this review, we present the most recent advances in diffusion models for vision, specifically focusing on the important design aspects that affect the computational efficiency of DMs. In particular, we emphasize the recently proposed design choices that have led to more efficient DMs. Unlike the other recent reviews, which discuss diffusion models from a broader  perspective, this survey is aimed at pushing this research direction forward by highlighting the design strategies in the literature that are resulting in practicable models for the  research community. We also provide a future outlook of diffusion models in the vision domain from their computational efficiency viewpoint.  
\end{abstract}
% === KEYWORDS ====================================================================
% =================================================================================

% ====================================================================

% \fi
\IEEEpeerreviewmaketitle

% ====================================================================
% ====================================================================
% ====================================================================

% === I. INTRODUCTION =============================================================
% =================================================================================
\section{Introduction}

\IEEEPARstart{D}{eep} generative modelling has emerged as one of the most exciting computational tools  that is even challenging human creativity~\cite{GMreview}. 

In the last decade, Generative Adversarial Networks (GANs)~\cite{brock2018large}, \cite{goodfellow2020generative} have received a lot of attention due to their high quality sample generation. However, diffusion models~\cite{cao2022survey,yang2022diffusion,croitoru2022diffusion} have recently emerged as an even more powerful generative technique, threatening the reign of GANs in the synthetic data generation.  

Diffusion models are becoming increasingly popular due to their superior training stability compared to GANs, and their ability to generate higher quality samples. They address several well-known limitations of GANs, such as mode collapse, the burden of adversarial learning, and issues with convergence~\cite{ganweakness}. 
Diffusion models employ a distinct training method from GANs, wherein the training data is contaminated with Gaussian noise, and the model learns to recover the original data from the noisy version. These models are advantageous in terms of scalability and parallelizability, making them even more appealing. Their training process applies  small modifications to the original data and rectifies those. This is used for learning the underlying data distribution that can generate samples with high degree of realism.
Diffusion models have been able to strongly influenced the current state-of-the-art in image generation, achieving remarkable results~\cite{faces,DALLE}.

\begin{figure*}
    \centering
    \includegraphics[width = \textwidth]{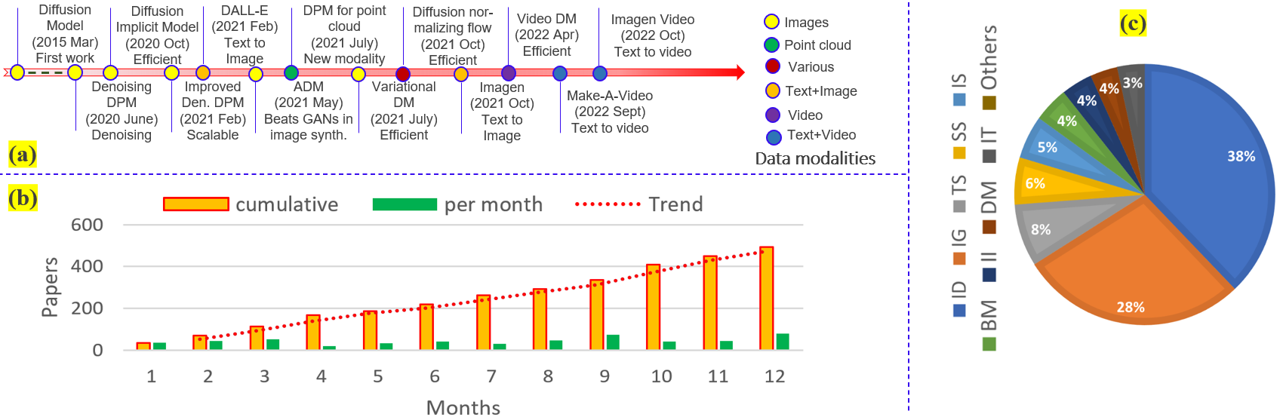}
    \caption{(\textbf{a}) Timeline of notable developments (non-exhaustive) in diffusion modelling. (\textbf{b}) The number of per-month and accumulative papers in diffusion models in the last 12 months based on Google Scholar search. (\textbf{c}) The proportion of research papers in terms of main application areas for diffusion models. The applications include Image Denoising (ID), Image Generation (IG), Time Series (TS), Semantic Segmentation (SS), Image Super-resolution (IS), BIG-bench Machine learning (BM), Image Inpainting (II), Decision Making (DM), and Image-to-image Translation (IT).}
    \label{fig:graphs}
\end{figure*}

Due to their amazing generative abilities, diffusion models are quickly finding applications in both low- and high-level vision tasks, including but not limited to image denoising~\cite{ho2020denoising}, \cite{nichol2021improved},  inpainting~\cite{esser2021imagebart}, image super-resolution~\cite{li2022srdiff}, \cite{batzolis2021conditional},  \cite{lugmayr2022repaint}, semantic segmentation~\cite{baranchuk2021label}, \cite{graikos2022diffusion}, \cite{wolleb2021diffusion},   image-to-image translation \cite{croitoru2022diffusion} etc. Hence, unsurprisingly, since the  seminal advancement of diffusion probabilistic models \cite{DDPM} over the original proposal of diffusion modelling~\cite{sohl2015deep}, there has been a continuous rise in the number of research papers appearing in this direction, and new exciting models are emerging everyday. In particular, diffusion modelling has gained a considerable social media hype after DALL-E~\cite{DALLE}, Imagen~\cite{Imagen}, and  Stable~\cite{rombach2022text} models that enabled high quality text-to-image generation. This hype has recently been fuelled further by the text-to-video generation techniques, where the videos appear considerably sophisticated~\cite{singer2022make}, \cite{ho2022imagen}. 
Figure~\ref{fig:graphs} provides statistics and a timeline overview of the recent literature on diffusion models to show their  popularity, particularly in the vision community.
%Fig. ~\ref{fig:graphs} provides statistics and a timeline overview of the literature on Diffusion Models to show their recent popularity in the vision community. 

% \begin{figure*}[t]
 %   \centering
 %   \includegraphics[width = 0.97\textwidth]{Graph1.png}
 %   \caption{Statistics and timeline overview of the literature on %Diffusion Models. (left) Number of per-month papers published in %diffision models in the last 12 months. (right) The proportion of %research papers in terms of application areas for diffusion models.}
%    \label{fig:graphs}
%\end{figure*}

%  \begin{figure*}[!tbp]
%  \centering
%   \begin{minipage}[b]{0.45\textwidth}
%  \includegraphics[width=\textwidth]{Papers.png}
%     \end{minipage}
%   \hfill
%  \begin{minipage}[b]{0.45\textwidth}
%   \includegraphics[width=\textwidth]{Tasks.png}
%  \end{minipage}
%     \caption{Statistics and timeline overview of the literature on Diffusion Models. (left) The number of per-month papers published in diffusion models in the last 12 months. (right) The proportion of research papers in terms of application areas for diffusion models. The applications include image denoising (ID), image generation (IG), time series (TS), semantic segmentation (SS), image super-resolution (IS), BIG-bench machine learning (BM), image inpainting (II), decision making (DM), and image-to-image translation (IT).}
%      \label{fig:graphs}
% \end{figure*}

Diffusion models belong to a category of probabilistic models that require excessive computational resources to model unobserved data details. Their training process requires evaluating models that follow iterative estimations and gradient computations. The computational cost becomes particularly huge while dealing with high dimensional data like images and videos~\cite{truncated}. For instance, a high-end diffusion model training in \cite{ADM}  takes 150-1000 V100 GPU days. Moreover, since the inference stage also  requires repeated evaluations of the noisy input space, this stage is also computationally demanding. In \cite{ADM}, 5 days of A100 GPU are required to produce 50k samples.  
Rombach et al.~\cite{rombach2022text} rightly noted that the huge computational requirements to train effective diffusion models present a critical bottleneck in terms of democratizing this technology because the research community generally lacks such resources. It is evident that the most exciting results using diffusion models are first achieved by e.g., Meta AI~\cite{singer2022make} and Google Research~\cite{ho2022imagen} who have an enormous computational power at their disposal. 
It is also notable that evaluating an already trained model has a considerable associated time and memory cost because the model may need to run for multiple steps (e.g., 25-1000) to generate a sample \cite{LDM}. This is a potential hindrance in the practical applications of diffusion models, especially in the resource constrained environments.

%This has two implications for the research community and general users: first, training such a model requires a lot of computing resources, is only suitable for a small part of the domain, and leaves a huge carbon footprint. Second, evaluating an already trained model is also expensive in terms of time and memory because the same model architecture needs to be run consecutively for a large number of steps (e.g. 25 - 1000 steps) \cite{LDM}.

In the contemporary era of large-scale data, early works on  diffusion models  focused on high-quality sample generation, largely disregarding the computational cost \cite{DDPM,ADM,san2021noise}. However, after achieving reasonable quality  milestones, the more recent works have also started to consider  computational efficiency, e.g.,~\cite{rombach2022text}, \cite{song2020denoising}, \cite{DiffFlow}. In particular, to address the genuine drawback of a slow generation process at the inference stage,  new trends are setting up, focusing on efficiency gains. 
In this review article, we collectively term the diffusion models evolved under the computational efficiency perspective, efficient diffusion models. These are the emerging models that are more valuable to the research community because they demand  accessible computational resources. Whereas progress is being consistently  made in terms of improving the computational efficiency, diffusion models are still far slower than GANs in terms of sample generation \cite{yang2022your,jing2022subspace}. We review the existing works concerned with  efficiency without sacrificing the high quality of sample generation. Moreover, we discuss the trade-offs between the model speed and sampling quality.

%Diffusion models rely on a long Markov chain of diffusion steps to generate samples, so it can be quite expensive in terms of time and computing. New methods have been proposed to make the process much faster, but the sampling is still slower than GAN \cite{yang2022your,jing2022subspace}.

\begin{figure*}[t]
    \centering
    \includegraphics[width = 0.9\textwidth]{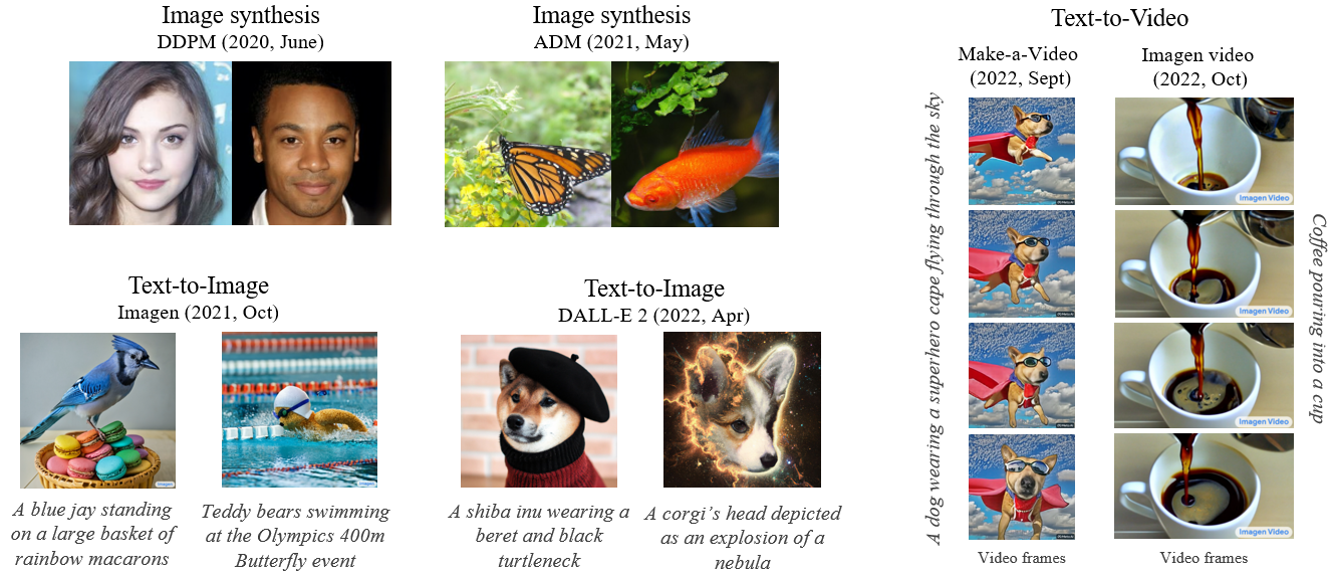}
    \caption{State-of-the-art diffusion models are able to generate excellent  quality samples for different tasks with minimal effort on their user's part. This portends a large-scale use of these models in the future in the applications ranging from research to entertainment. The shown images are cropped from the original works.}
    \label{fig:examples}
\end{figure*}

\vspace{0.7mm}
\noindent\textbf{Why model efficiency is critical?} Diffusion models have been able to produce an astonishing quality of images and videos, virtually requiring no effort on their users' part - see Fig.~\ref{fig:examples}. This foretells a widespread use of these  models in the daily-life applications in the future. However, the creative abilities of diffusion models do not come for free. High-quality generative modeling is energy-intensive.
%, and the higher the quality demand, the more power it consumes. 
Training a sophisticated AI model needs time, money, and power \cite{carbonai, powerloss}, leaving behind a significant carbon footprint. 
%Greenhouse gases, such as CO2, trap heat near the Earth's surface in the atmosphere, raising global temperatures and upsetting fragile ecosystems. 
To put things into a perspective, OpenAI trained GPT-3 model \cite{GPT3} On 45 terabytes of data. NVIDIA trained the final version of MegatronLM, a language model comparable to but smaller than GPT-3, using 512 V100 GPUs for nine days. A single V100 GPU may consume up to 300 Watts of power. If we estimate the power consumption with a conservative estimate of 250 Watts, 512 V100 GPUs utilise 128,000 Watts or 128 kilo-Watts (kW) \cite{v100}. Running for nine days requires 27,648 kWh of energy for the MegatronLM. The average home consumes 10,649 kWh per year as per the US Energy Information Administration. Implying, training MegatronLM required nearly as much energy as three houses use in a year. Among the currently most hyped diffusion models (due to their ability to perform text-to-image task), e.g., DALL-E~\cite{DALLE}, Imagen~\cite{Imagen}, and  Stable~\cite{rombach2022text}, Stable is by far the most efficient because its diffusion process is mainly carried out in a lower dimension latent space. However, even the training of this model requires an energy equivalent to burning nearly 7,000 kgs of coal\footnote{Computed with \url{https://mlco2.github.io/impact/#compute}}. Not to mention, the text-to-image diffusion models already rely on language models such as GPT-3 mentioned above. Other diffusion models, especially for the more complex tasks, e.g.~text-to-video, are expected to require orders of magnitude more energy\footnote{The carbon footprint calculations are not possible from the details provided in the original papers.}. Hence, due to the fast growing popularity of these models, it is critical to focus on more efficient schemes.

\vspace{0.7mm}
\noindent\textbf{Motivation and uniqueness of this survey}: Since  the diffusion models have recently received a significant attention of the research community, the literature is experiencing a large influx of contributions in this direction. This has also led to  review articles surfacing recently. Among them, Yang et al.~\cite{yang2022diffusion} reviewed the broad direction of diffusion modelling from the methods and applications viewpoint, and Cao et al.~\cite{cao2022survey} also discussed diffusion models more broadly. More related to our review is \cite{croitoru2022diffusion}, which focuses on the diffusion model in the vision domain. On one hand, all these reviews already surfaced before this direction fully matured. For instance, the breakthrough of high quality text-to-video generation with diffusion models~\cite{singer2022make}, \cite{ho2022imagen} is actually achieved after the appearance of all these surveys. On the other hand, none of these surveys focuses on computational efficiency of the models, which is the central aspect in pushing this research direction forward. Hence, these surveys leave a clear open gap. We aim at addressing that by  highlighting the underlying schemes of the techniques that are improving the computational efficiency of diffusion models. Our comprehensive review of the existing methods from this pragmatic perspective is expected to advance this research direction in the ways not covered by the reviews appeared during the preparation of this article. %\footnote{We note that this manuscript is still a work in progress, which will be updated in the future by improving its quality and including further progress in this direction.}. 

%The motivation of this review paper is to explore the design of Diffusion approaches in-depth and highlight design choices that can provide insight into corrections Model efficiency. In contrast to the previous works that classify diffusion models in general, this review article will present a precise classification of the design choices that lead to effective and ineffective diffusion models. This will guide future research on computationally efficient diffusion models for computer vision tasks.

The rest of the article is organised as follows: Section~\ref{sec:Overview} provides an overview of diffusion models with a brief discussion on three representative architectures.  Section~\ref{sec:Strategies} provides a description of design choices and discusses how these choices lead to computation-efficient designs. In Section~\ref{sec:Compare} compares representative works w.r.t quality and efficiency trade-off. Section~\ref{sec:Conc} discusses future work directions, followed by a conclusion.

% === II. Harmonically-Terminated Power Rectifier Analysis ========================
% =================================================================================
\section{An Overview of Diffusion  Models}
\label{sec:Overview}
Probabilistic diffusion models are designed to learn how to invert a process that progressively corrupts the structure of the training data by introducing noise at varying levels of granularity.
%The original idea of the probabilistic diffusion model is to model a specific distribution from random noise. Therefore, the distributions of the generated samples should be as close as those of the original samples. 
Their training process includes a forward process (or diffusion process), in which complex data (generally an image) is progressively corrupted, and a reverse process (or reverse diffusion process), in which noise is transformed back into a sample from the target distribution.
Here, we describe three models in particular due to their high influence on efficient diffusion modelling. It includes Denoising Diffusion Probabilistic Model (DDPM) \cite{DDPM}, Latent Diffusion Model (LDM) \cite{LDM} and Feature Pyramid Latent Diffusion Model \cite{Frido}.

\subsection { The Baseline: Denoising Diffusion Probabilistic Model (DDPM):}

Here's the corrected version with the formula and LaTeX symbols:

Suppose we have an original data point sampled from a real data distribution $\mathbf{x}_0 \sim q(\mathbf{x})$. We can define a forward diffusion process where we gradually add a small amount of Gaussian noise to the samples, resulting in a series of noisy samples $\mathbf{x}_1, \dots, \mathbf{x}T$. The step size of the process of noise addition is controlled by the variance schedule ${\beta_t \in (0, 1)}{t=1}^T$ of the distribution. Mathematically,

\begin{equation}\label{1}
q(\mathbf{x}t \vert \mathbf{x}{t-1}) = \mathcal{N}(\mathbf{x}t; \sqrt{1 - \beta_t} \mathbf{x}{t-1}, \beta_t\mathbf{I}), \
q(\mathbf{x}_{1:T} \vert \mathbf{x}0) = \prod{t=1}^T q(\mathbf{x}t \vert \mathbf{x}{t-1})
\end{equation}

The actual learning process of a diffusion model occurs in the reverse direction. The model is trained to denoise the progressively corrupted samples. Here, the model is a neural network that approximates the conditional probabilities to reverse the diffusion process.

\begin{equation}\label{2}
p_\theta(\mathbf{x}{0:T}) = p(\mathbf{x}T) \prod{t=1}^T p\theta(\mathbf{x}{t-1} \vert \mathbf{x}t), \
p\theta(\mathbf{x}{t-1} \vert \mathbf{x}t) = \mathcal{N}(\mathbf{x}{t-1}; \boldsymbol{\mu}_\theta(\mathbf{x}t, t), \boldsymbol{\Sigma}\theta(\mathbf{x}_t, t))
\end{equation}

In the above equation, $\theta$ denotes the neural network's parameters. Sohl-Dickstein et al.~\cite{sohl2015deep} has noted that when $\beta_t$ in the variance schedule is small enough, then the conditional probability $p_\theta(\mathbf{x}_{t-1} \vert \mathbf{x}_t)$ is well-approximated as a Gaussian distribution. This entails that the underlying neural network can directly learn to predict the mean and variance parameters of this distribution. In DDPM, the forward or diffusion process imitates a Markov chain that follows the variance schedule to gradually add Gaussian noise to the data. Hence, the reverse Markov transitions that maximize the probability of the training data are used to train the diffusion model. The model training in practice is similar to reducing the variational upper bound on the negative log probability. Because this is similar to VAE, we may apply the variational lower bound to optimize the negative log-likelihood.

Here's the corrected version with the proper formatting for the formula and LaTeX symbols:

\begin{equation}\label{4}
\text{Let } L_\text{VLB} = \mathbb{E}{q(\mathbf{x}{0:T})} \Bigg[ \log \frac{q(\mathbf{x}{1:T}|\mathbf{x}0)}{p\theta(\mathbf{x}{0:T})} \Bigg] \geq -\mathbb{E}_{q(\mathbf{x}0)} \log p\theta(\mathbf{x}_0)
\end{equation}

To make each component in the equation analytically computable, the objective may be reformulated as a mixture of many KL-divergence and entropy terms. Let us label each component of the variational lower bound loss separately.

\begin{figure*}[t]
   \centering
   \includegraphics[width = 0.97\textwidth]{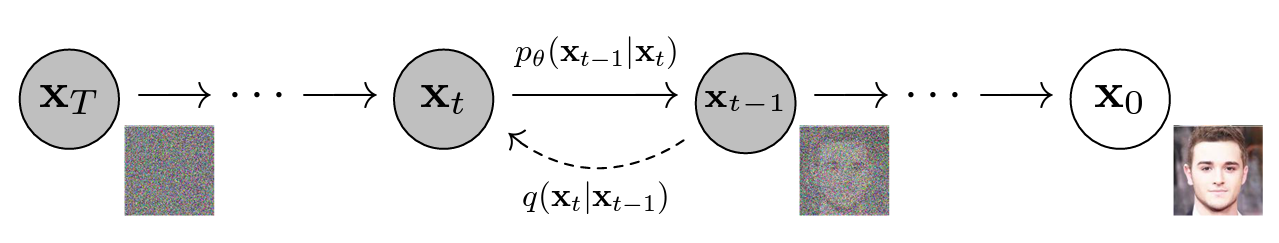}
   \caption{The directed graphical model illustrates processes involved in a diffusion model. The original sample $\boldsymbol{x}_0$ gets gradually corrupted with a Markov process to look like noise $\boldsymbol{x}_T$. The model learns to denoise the corrupted image at every step by learning the conditional probability $p_{\theta} (\boldsymbol{x}_{t-1}| \boldsymbol{x}_t)$. Image taken from~\cite{DDPM}.}
      \label{fig:DDPM}
\end{figure*}

\begin{equation}\label{5}
\begin{aligned}
L_\text{VLB} &= L_T + L_{T-1} + \dots + L_0 \\
\text{where } L_T &= D_\text{KL}(q(\mathbf{x}T \vert \mathbf{x}0) \parallel p\theta(\mathbf{x}T)) \\
L_t &= D\text{KL}(q(\mathbf{x}t \vert \mathbf{x}{t+1}, \mathbf{x}0) \parallel p\theta(\mathbf{x}t \vert \mathbf{x}{t+1})) \text{ for } 1 \leq t \leq T-1 \\
L_0 &= -\log p\theta(\mathbf{x}_0 \vert \mathbf{x}_1)
\end{aligned}
\end{equation}

%%%%%%%%%%%%%%%%%%%%%%%%%%%%%%%%%%%%%%%%%%%

Since every KL term in $L_\text{VLB}$ (excluding $L_0$) compares two Gaussian distributions, they can be calculated in closed form. In the reverse diffusion process, a neural network is trained to approximate the conditioned probability distributions.

As $\mathbf{x}_t$ is available as input at the training time, the Gaussian noise term can be reparameterized as

\begin{equation}\label{6}
\text{Thus } \mathbf{x}{t-1} = \mathcal{N}(\mathbf{x}{t-1}; \frac{1}{\sqrt{\alpha_t}} \Big( \mathbf{x}_t - \frac{1 - \alpha_t}{\sqrt{1 - \bar{\alpha}t}} \boldsymbol{\epsilon}\theta(\mathbf{x}t, t) \Big), \boldsymbol{\Sigma}\theta(\mathbf{x}_t, t))
\end{equation}

Empirically, training the diffusion model works better with a simplified objective that ignores the weighting term.

\begin{equation}\label{7}
L_t = \mathbb{E}_{t \sim [1, T], \mathbf{x}_0, \boldsymbol{\epsilon}_t} \Big[|\boldsymbol{\epsilon}t - \boldsymbol{\epsilon}\theta(\mathbf{x}t, t)|^2 \Big] \
= \mathbb{E}{t \sim [1, T], \mathbf{x}_0, \boldsymbol{\epsilon}_t} \Big[|\boldsymbol{\epsilon}t - \boldsymbol{\epsilon}\theta(\sqrt{\bar{\alpha}_t}\mathbf{x}_0 + \sqrt{1 - \bar{\alpha}_t}\boldsymbol{\epsilon}_t, t)|^2 \Big]
\end{equation}

The final simple objective is $L = L_t + C$, where $C$ is a constant not depending on $\theta$.

\vspace{1mm}

\noindent\textbf{Model Efficiency}: Leaving aside the training process, it is very time consuming to generate a sample from DDPM by following the Markov chain of the reverse diffusion process, as $T$ can be up to a few thousand steps. For example, it takes around 20 hours to sample 50k images of size $32 \times 32$ from a DDPM using an NVIDIA 2080 Ti GPU. To put this number into a perspective, it takes less than a minute to do so from a GAN using the same hardware specifications.

\subsection {Latent Diffusion Model (LDM):}
A defining attribute of Latent Diffusion Model (LDM)~\cite{LDM} is that it performs the diffusion process in the latent space rather than the pixel space. This lowers the training costs and and also improves the inference speed.  
The central idea is driven by the fact that majority of the image bits only contribute to refine the perceptual details and that the semantic and conceptual composition of the image  persists even after extreme compression. With generative modelling, the LDM loosely decomposes perceptual compression and semantic compression by first cutting out the pixel-level redundancy with autoencoder and then manipulating/generating semantic ideas with the diffusion process on the learnt latent.

An autoencoder model, say $\mathcal{E}$, is used in the perceptual compression process. It compresses the input image $\mathbf{x} \in \mathbb{R}^{H \times W \times 3}$ to a smaller 2D latent vector

$\mathbf{z} = \mathcal{E}(\mathbf{x}) \in \mathbb{R}^{h \times w \times c}$ using the downsampling rate $f=H/h=W/w=2^m, m \in \mathbb{N}$. Then, a decoder $\mathcal{D}$ reconstructs the image from the latent vector as $\tilde{\mathbf{x}} = \mathcal{D}(\mathbf{z})$.

The neural backbone of the LDM is realized as a time-conditional UNet. The model has the ability to build the underlying UNet primarily from 2D convolutional layers and further focus the objective on the perceptually most relevant bits using the reweighted bound, which reads for LDM as

\begin{equation}\label{8}
L_\text{LDM} = \mathbb{E}_{t \sim [1, T], \mathbf{x}_0, \boldsymbol{\epsilon}_t} \Big[|\boldsymbol{\epsilon}t - \boldsymbol{\epsilon}\theta(\mathbf{z}_t, t)|^2 \Big]
\end{equation}

\begin{figure*}[t]
   \centering
   \includegraphics[width = 0.7\textwidth]{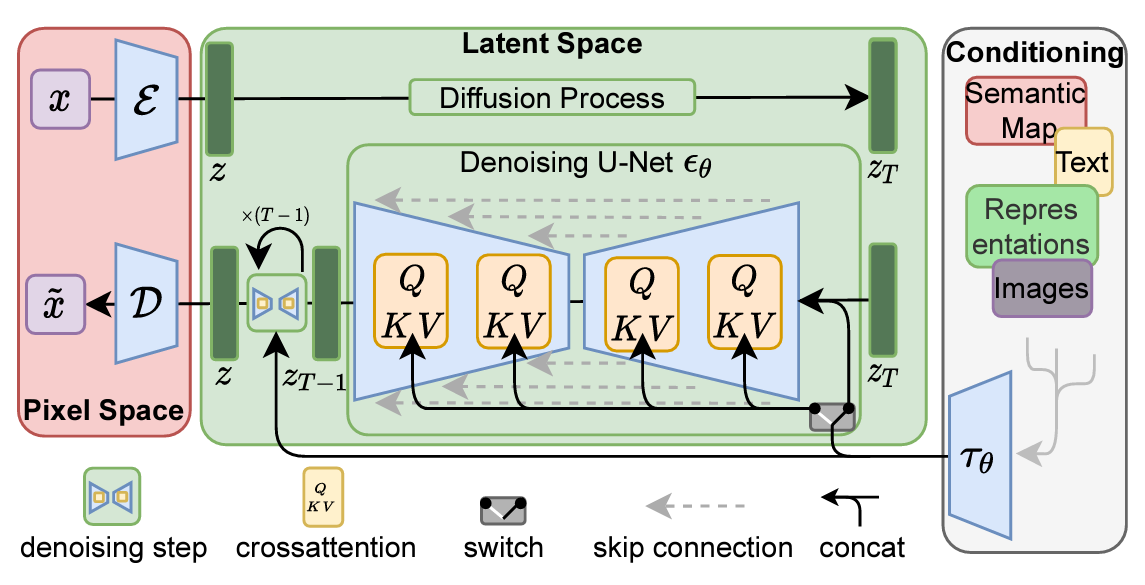}
   \caption{The architecture of the latent diffusion model (LDM) that is considered a revolutionary work that has been employed in stable diffusion and turned the direction of research towards efficient discussion models in general. (Source:\cite{LDM} )}
    \label{fig:LDM}
\end{figure*}

For LDM, the diffusion and denoising processes take place on the latent vector $\mathbf{z}$. The denoising model is a conditioned U-Net that is supplemented with a cross-attention mechanism to manage flexible conditioning information for image production (e.g., class labels, semantic maps, blurred variants of an image). The design is akin to fusing representations of various modalities into a model with a cross-attention mechanism. Each kind of conditioning information is associated with a domain-specific encoder

$\tau_\theta$, which projects the conditioning input to an intermediate representation $\tau_\theta(y) \in \mathbb{R}^{M \times d_\tau}$ that can be mapped to the intermediate UNet layers using cross-attention implemented as

\begin{equation}\label{9}
\text{Attention}(\mathbf{Q}, \mathbf{K}, \mathbf{V}) = \text{softmax}\Big(\frac{\mathbf{Q}\mathbf{K}^\top}{\sqrt{d}}\Big) \cdot \mathbf{V}
\end{equation}

where,

\begin{equation}\label{10}
\mathbf{Q} = \mathbf{W}_Q^{(i)} \cdot \varphi_i(\mathbf{z}_i), \quad
\mathbf{K} = \mathbf{W}K^{(i)} \cdot \tau\theta(y), \quad
\mathbf{V} = \mathbf{W}V^{(i)} \cdot \tau\theta(y)
\end{equation}

In the above, $\varphi_i(\mathbf{z}i) \in \mathbb{R}^{N \times d\epsilon^i}$ is a flattened representation and

\begin{equation}\label{11}
\mathbf{W}Q^{(i)} \in \mathbb{R}^{d \times d\epsilon^i}, \quad
\mathbf{W}_K^{(i)}, \mathbf{W}V^{(i)} \in \mathbb{R}^{d \times d\tau}
\end{equation}

are learnable weight matrices

%%%%%%%%%%%%%%%%%%%%%%%%%%%%%%%%%
\begin{figure*}[t]
   \centering
   \includegraphics[width = 0.8\textwidth]{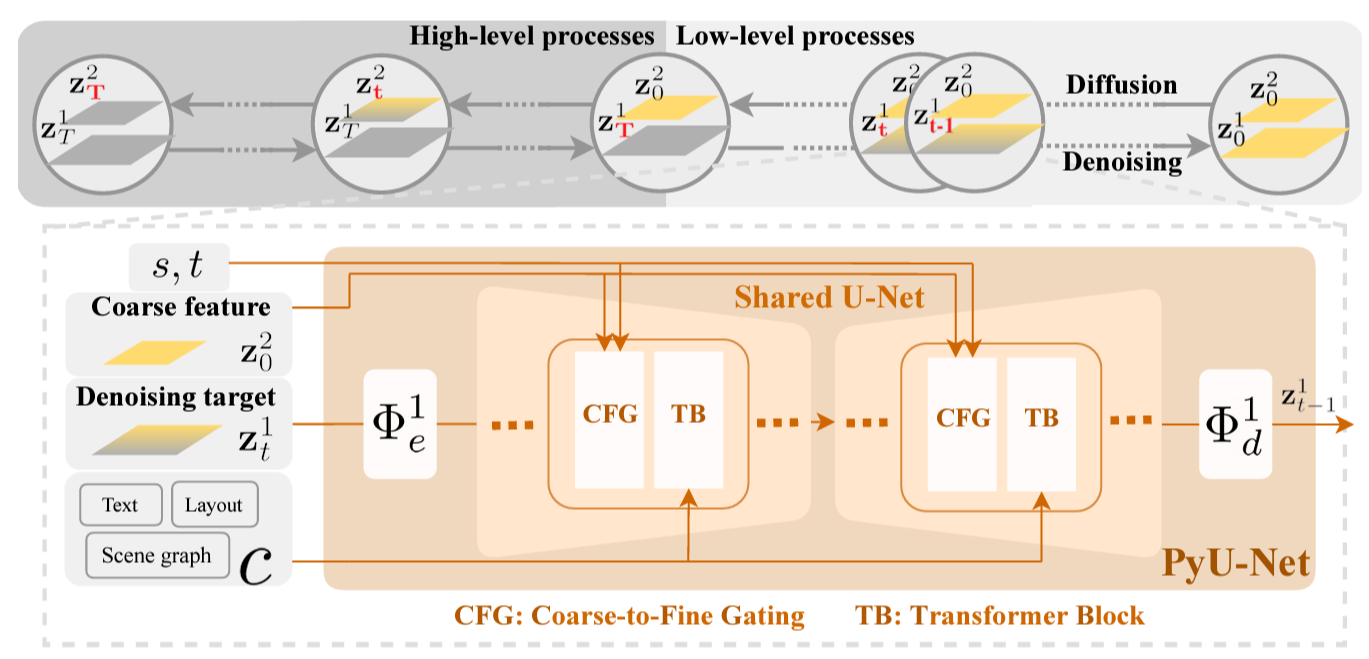}
   \caption{The architecture of the Feature Pyramid Diffusion Model (Frido) encodes an image into multi-scale feature maps $\mathbf z$ to improve the efficiency of diffusion models.  (Source:\cite{Frido})}
    \label{fig:Frido}
\end{figure*}

%%%%%%%%%%%%%%%%%%%%%%

\subsection {Feature Pyramid Latent Diffusion Model (Frido):}
 
Frido~\cite{Frido} decomposes the input image into scale-independent quantized features and then obtains the output result through a coarse-to-fine gating. In short, this techniques first uses multi-scale MS-VQGAN~\cite{MSVQGAN} to  encode the input image into a latent space, and then uses a feature pyramid UNet to perform denoising in the latent space.
The MS-VQGAN encodes the input image into N-scale latent variables, similar to an image pyramid. However, it is done  in the latent space. The low-level latent variables retain the lower-level visual details, while the high-level latent variables are able to maintain  the high-level shapes and structures. The decoder decodes the obtained hidden variables of all scales into the output image. The size of the pyramid of this hidden variable also decreases with the number of layers, and each layer size is half of the previous layer size. Overall, this leads to maintaining both high-level semantic information and  low-level details. 

% Need correction

Given an image $x_0$, the encoder $\mathcal E$ first produces a latent feature map set of $N$ scales

$$
Z = {\mathcal{E}} (x_0) = \left\{ \mathbf{z}^1, \mathbf{z}^2, \dots, \mathbf{z}^N \right\}
$$

Next, comes the diffusion process in the latent space. This process is  sequentially performed on different scales, and for each scale, $T$ step are required. Then, the diffusion operation of adding noise first destroys the details of the image, then the high-level shape, and finally the structure of the entire image.

Similar to the denoising process, information progressively flows from high-resolution (coarse) features to low-resolution (fine) features. Building upon the widely known U-Net architecture, the authors propose a feature pyramid UNet (PyU-Net) \cite{Frido} to achieve multi-scale denoising.

Coarse-to-fine gating is added to allow low-level denoising to utilize existing high-level information. For more effective training, it uses a teacher forcing trick, which  maintains the  training efficiency while preventing overfitting, and enables UNet to obtain information on the current scale level and time step. Finally, another level-specific projection decodes the UNet output to predict the noise added on the latent $\mathbf z$ with the following objective.

\begin{equation}
\label{13}
L_{Frido} = \mathbb{E}_{z_0^n, \epsilon, t} \left[ ||\epsilon - \epsilon_\theta (\mathbf{z}_t^n, \mathbf{z}_t^{n+1: N}, t)||^2 \right].
\end{equation}

%%%%%%%%%%%%%%%%%%%%%%
%%%%%%%%%%%%%%%%%%%%%%%%%%%%%%%%%%%%%
\section {Strategies for Efficient  Diffusion Models }
\label{sec:Strategies}
Diffusion models learn the underlying  distribution of the training data. Hence, their generation process requires distribution sampling. One of the major shortcomings of the diffusion models is that their sampling process is inefficient, which renders data  generation very slow.
%to generate samples from DDPM. 
Diffusion models must rely on a long Markov chain of the diffusion steps to generate samples, which is expensive in terms of both time and computing resources. 
The current literature has witnessed multiple efforts to accelerate the data generation with diffusion models. We divide the major strategies in that regard into two broad categories, namely; Efficient Design Strategies (EDS) and Efficient Process Strategies (EPS). The former recommend modifications to the design of the baseline diffusion model for efficiency, whereas the latter suggest ways to improve the efficiency of the diffusion models or speed up the sampling process.
We infer these strategies based on the existing literature. It is anticipated that  further strategies will  emerge in the future research due to the criticality of the issue.

\begin{figure*}[t]
   \centering
   \includegraphics[width = 0.6\textwidth]{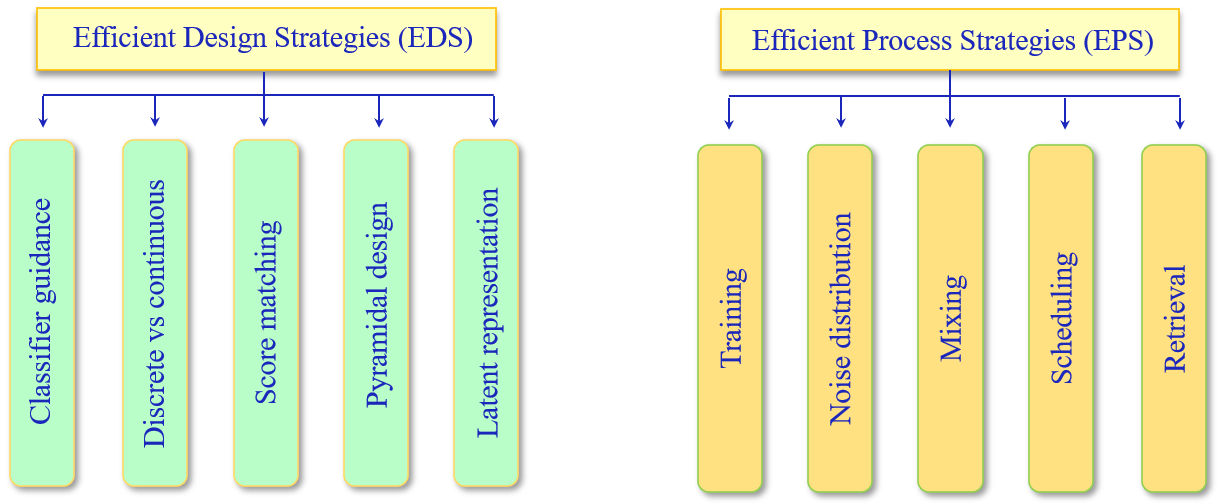}
   \caption{The influencing strategies for efficient diffusion models can be divided into two categories: Efficient Design Strategies (EDS), which recommend modifications to the design of baseline diffusion models, and Efficient Process Strategies (EPS), which suggest ways to improve the efficiency of diffusion models or speed up the sampling process.}
    \label{fig:Strategies}
\end{figure*} 

\subsection {Efficient Design Strategies (EDS)}
  These strategies  mainly pertain to  the architectural design of the diffusion models.
  We provide a brief description of each strategy with the help of representative works from the literature below. A summary of the representative works from the reviewed literature is also provided in  
  Table~\ref{tab:EDS}.
  %includes some representative works in each architectural category that is included. A brief description of each category and its influence on the efficiency of diffusion models are discussed below: 

%%%%%%%%%%%%%%%%%%%%%%%%%%%%%%%%%%%%%%%%%%%%%%%%%%%%%%
\begin{table}
%\begin{center}
\centering
 \begin{tabular}{|l|l|l|p{6cm}|}
\hline
Architecture & Model        & Citation & Strategy                            \\
\hline
Guided       & ADM          &  \cite{ADM}        & Classifier guidance and  upsampling \\
\hline
Not-Guided   & CFDG         &   \cite{CFDG}       & Classifier-free guidance            \\
\hline
Guided       & SDEDIT       &    \cite{SDEDIT}      & Stochastic Differential Editing     \\
\hline
Guided       & VDM          &    \cite{VDM}       & Reconstruction-guided sampling      \\
\hline
Guided       & SDG          &   \cite{SDG}        & Semantic Diffusion Guidance         \\
\hline
Not-Guided   & Make-A-Scene &   \cite{Make-A-Scene}        & Classifier-free guidance            \\
\hline
Not-Guided   & VQ-Diffusion &   \cite{VQ-Diffusion}        & DiscreteClassifier-free Guidance   \\
\hline
Discrete     & DDPM        &  \cite{DDPM}        & Discrete data                              \\
\hline
Continuous   & DP stride    &  \cite{DPstride}        & Continuous Time Affine Diffusion Processes \\
\hline
Continuous   & FastDPM      &    \cite{FastDPM}      & Continuous diffusion steps                 \\
\hline
Continuous   & DSB          &   \cite{DSB}       & Diffusion Schrödinger Bridge               \\
\hline
Discrete     & VQ-Diffusion & \cite{VQ-Diffusion} & Discrete Classifier-free Guidance \\  \hline  

Score Network & BDDM         &     \cite{BDDM}           & score and scedulng network      \\
\hline
Score Network & CLD          &      \cite{CLD}          & Score matching objective        \\
\hline
Score Network & GGDM         &      \cite{GGDM}          & Sample quality scores.          \\
\hline
SDE           & DSB          &    \cite{DSB}            & Diffusion Schrödinger Bridge    \\
\hline
Score Network & ScoreFlow & \cite{ScoreFlow} & Inproving  likelihood of Scores \\
\hline
SDE           & EMSDE        &      \cite{EMSDE}          & Adaptive step sizes  \\
\hline
Pyramidal     & CDM            &\cite{CDM}  & Cascading pipelines                    \\
\hline
Pyramidal     & Frido          & \cite{Frido} & Pyramid diffusion                      \\
\hline
Pyramidal     & PDDPM          & \cite{PDDPM} & Pyramidal reverse sampling             \\
\hline
Non-Pyramidal & Blur diffusion & \cite{Blurr} & Frequency Diffusion at variable speeds \\
\hline
Non-Pyramidal & NWDM           & \cite{NWDM} & Frequency domain diffusion  \\ 
\hline
Latent & ILVR        & \cite{ILVR} & Iterative Latent Variable Refinement \\
\hline
Latent & LDM         & \cite{LDM} & latent space                         \\
\hline
Latent & latent DDIM & \cite{LDDIM} & Semantic latent interpolation        \\
\hline
Latent & INDM        & \cite{INDM} & Linear diffusion on the latent space \\
\hline
Pixel  & DDPM        & \cite{DDPM} & image upsampling and downsampling    \\
\hline
Pixel   & DDIM        &\cite{DDIM}  & image upsampling and downsampling \\  
\hline
\end{tabular}
%\captionsetup{singlelinecheck=false, font=small, labelfont=bf}
   \caption{Representative works on Efficient design strategies (EDS) with mention of the model, architectural approach, citation and strategy used in the existing literature.  These strategies are based on the architecture of diffusion models. }
  %\end{center}
   \label{tab:EDS}
    \end{table}

%%%%%%%%%%%%%%%%%%%%%%%%%%%%%%%%%%%%%%%%%%%%%%%%%%%%%%

\subsubsection{Classifier Guidance} Classifier guidance is a recently developed strategy for balancing mode coverage and sample fidelity in post-training conditional diffusion models. It works  in the same way that low-temperature sampling or truncation is used in other forms of generative models. An example is a work by  Nichol et al.~\cite{glide}, where initially, the classifier undergoes training using images that have been corrupted by noise. Subsequently, when performing the diffusion sampling process, the classifier's gradients are utilized to steer the sample towards the intended classification label.

Classifier guidance provides a trade-off. On one hand, it enhances adherence to the conditioning signal and improves the overall sample quality. On the other, it can also incur high computational cost for high quality samples. The efficiency aspect of classifier guidance is better understood when the computational requirements of sample generation for a specific objective are considered in the absence of any  guidance.  While classifier guidance help improving the quality metrics, as expected from truncation or low-temperature sampling, it is nonetheless reliant on the gradients from an image classifier.
Addressing that, the classifier-free guidance \cite{CFDG} achieves the same effect without relying on such gradients. Classifier-free guidance is an alternative method of modifying gradients but without a classifier. However, it does come at the cost of reducing the sample diversity for the diffusion model.
  
\subsubsection{Discrete vs Continuous Design} Diffusion is a continuous process that can be characterized by a stochastic differential equation. The probability flow ODE (diffusion ODE) is a continuous-time differential equation \cite{vahdat2021score}. Denoising diffusion probabilistic models (DDPMs) \cite{DDPM} have shown impressive results in image and waveform generation in continuous state spaces.

Denoising diffusion models have produced both remarkable log-likelihood scores on numerous common picture datasets and high-quality image production in continuous situations. Many datasets are discrete, but for the convenience of modeling, they are frequently embedded in a continuous space and modeled continuously.

Structured corruption processes appropriate for text data, using similarity between tokens to enable gradual corruption and denoising. Diffusion models with discrete state spaces were first introduced by Sohl-Dickstein et al.\cite{sohl2015deep} who considered a diffusion process over binary random variables. Considered a simple 2×2 transition matrix for binary random variables. Hoogeboom et al. \cite{ARDMs} later extended this to categorical variables, proposing a transition matrix.

This, however, can lead to tough modeling concerns such as "de-quantization" blockages, weird gradient issues, and difficulties understanding log-likelihood metrics. All of these concerns are avoided by representing discrete data separately. Instead of transitioning uniformly to any other state, for ordinal data, discrete models imitate a continuous space diffusion model by using a discretized, truncated Gaussian distribution. 

In terms of efficient design, discrete diffusion design is preferable as it helps reduce the number of samples. Diffusion models with discrete state spaces were first introduced by Sohl-Dickstein et al. \cite{sohl2015deep}, who considered a diffusion process over binary random variables.

Even though diffusion models have been presented in both discrete and continuous state spaces, much current work has concentrated on Gaussian diffusion processes that operate in continuous state spaces (e.g. for real-valued image and waveform data).

\subsubsection{Score Matching Networks or SDEs Design} The score network may be used to create an ODE ("scoring-based diffusion ODE") for evaluating precise probability \cite{CLD, SDB}. They simulate the distribution of data by matching a parameterized score network with first-order data score functions. The gradient of the log-likelihood concerning the random variable x is defined as the score.

\begin{equation}
    \nabla_\mathbf{x} \log p_\theta (\mathbf{x}) = - \nabla_\mathbf{x} f_\theta (\mathbf{x}).
\end{equation}

The purpose of score-matching is to reduce the difference between $p textdata$ and $p textdata$ by optimizing the Fisher divergence. It has been used in medical applications such as low-dose computed tomography (LDCT), resulting in a low signal-to-noise ratio (SNR) and potential impairment of diagnostic performance. The Diffusion Probability Model for Conditional Noise Reduction (DDPM) has been shown to improve LDCT noise reduction performance with encouraging results at high computational efficiency. Especially considering the high sampling cost of the original DDPM model, the fast ordinary differential equation (ODE) solver can be scaled for greatly improved sampling efficiency. Experiments \cite{xia2022low} show that accelerated DDPM can achieve 20X speedup without degrading image quality.

%%%%%%%%%%%%%%%%%%%%%%%%%%%%%%%%%%%%%%%%%%%%%%%%%%%%%%
\begin{table*}
\begin{center}
 \begin{tabular}{|l|l|l|p{5cm}|}
\hline
Process & Model        & Citation & Strategy                            \\
\hline
Training & FSDM & \cite{FSDM} & Conditioned on a small class set       \\ \hline
Training & VDM  & \cite{VDM}  & Joint training from image and video    \\ \hline
Training & MMDM & \cite{MMDM} & Argmax Flows and Multinomial Diffusion \\ \hline
Training & P2   & \cite{P2} & Perception PrioritizedWeighting        \\ \hline
Training & DDRM & \cite{DDRM} & Pre-trained demonising diffusion  \\ \hline
Noise Ditribution & DGM            & \cite{DGM} & noise from Gamma distribution           \\ \hline
Noise Ditribution & GGDM           & \cite{GGDM} & Differentiable Diffusion Sampler Search \\\hline
Noise Ditribution & NEDM           & \cite{NEDM} & Noise Schedule Adjustments              \\ \hline
Noise Ditribution & CCDF           &\cite{CCDF}  & Non Gaussian initialisation            \\ \hline
Noise Ditribution & Cold Diffusion & \cite{ColdDiffusion} & Deterministic noise degradation         \\ \hline
Noise Ditribution & BDDM            & \cite{BDDM} & Non-isotropic noise                     \\ \hline
Noise Ditribution & GDDIM          &\cite{GDDIM}  & Mixture Gaussian, Gamma Distribution  \\ \hline
Mixing & DiVAE         & \cite{DiVAE} & Input image embedding              \\ \hline
Mixing & FastDPM       & \cite{FastDPM} & Unified framework                   \\ \hline
Mixing & DiffFlow      & \cite{DiffFlow} & Normalizing flow and diffusion     \\ \hline
Mixing & MMDM          &\cite{MMDM}  & Argmax Flows Multinomial Diffusion \\ \hline
Mixing & LDM+DDIM      & \cite{LDM3} & Latent Implicit Diffusion          \\ \hline
Mixing & Diffusion-GAN & \cite{Diffusion-GAN} & Gaussian mixture distribution \\ \hline
Scheduling & DP stride      & \cite{DPstride} & Learning time schedule by optimization     \\ \hline
Scheduling & Bit Diffusion  & \cite{Bit} & Asymmetric Time Intervals                  \\ \hline
Scheduling & DiffusionTimes & \cite{DiffusionTimes} & Trade off on diffusion time                 \\ \hline
Scheduling & ProgressiveDistillation &\cite{ProgressiveDistillation} & Deterministic diffusion sampler \\ \hline
Scheduling & ES-DDPM        & \cite{ES-DDPM}  & Early Diffusion Stoppage                   \\ \hline
Scheduling & CMDE           &\cite{CMDE}  & Multiscale diffusion                       \\ \hline
Scheduling & FHDM           & \cite{FHDM} & Termination at a random first hitting time \\ \hline
Scheduling & IDSD           & \cite{IDSD} & Stochastic sampling  \\\hline
Retreival & KNN-Diffusion & \cite{KNN-Diffusion} & KNN adapted training              \\\hline
Retreival & RDM           &\cite{RDM}  & Database subset conditioning     \\ \hline
Retreival & RDM           &\cite{RDM2} & Informative samples Conditioning \\
\hline
\end{tabular}
\captionsetup{singlelinecheck=false, font=small, labelfont=bf}
   \caption{Representative works on Efficient process strategies (EPS) with mention of the model, process, citation and the strategy used in the existing literature. These strategies target the improvement of the diffusion process itself. }
  \end{center}
    \end{table*}

%%%%%%%%%%%%%%%%%%%%%%%%%%%%%%%%%%%%%%%%%%%%%%%%%%%%%%

A stochastic differential equation (SDE) \cite{SDEDIT} is a differential equation where one or more of the terms is a stochastic process, resulting in a solution, which is itself a stochastic process.

Diffusion ODE can be seen as a semi-linear form by which the discretization errors are reduced. DPM-solver accomplished the SOTA within 50 steps on CIFAR-10 \cite{CIFAR10}, and it can generate high-quality images) with ten steps, which is an extensive upgrade.

Compared to traditional diffusion methods with discrete steps, numerical formulations of differential equations achieve more efficient sampling with advanced solvers. Inspired by Score SDE and Probability Flow (Diffusion), ODE.

\subsubsection{Pyramidal or non-Pyramidal Design} Pyramidal approaches to training the diffusion model such that it can understand the different scales of the input by giving coordinate information as a condition. These models concatenate an input image and coordinate the values of each pixel. Then, random resizing to the target resolution is applied to the merged input. The resized coordinate values are encoded with the sinusoidal wave, expanded to high dimensional space, and act as conditions when training. Benefiting from the UNet-like model structure \cite{DiVAE}, the cost function is kelp invariant to all different resolutions so that the optimization can be performed with only a single network. The multi-scale score function, the sampling speed, which is the most critical disadvantage of the diffusion models, can also be made much faster compared to a single full DDPM by a reverse sampling process.

Therefore, the pyramidal or multiscale approach provides better efficiency to diffusion models.

\subsubsection{Pixel or Latent Representation-based Design} The majority of a digital image's bits correspond to insignificant information. While DMs allow for the suppression of semantically meaningless information by minimizing the responsible loss term, gradients (during training) and the neural network backbone (during training and inference) must still be evaluated on all pixels, resulting in redundant computations and unnecessarily expensive optimization and inference.

The model class Latent Diffusion Models (LDMs) provide efficient image generation from the latent space with a single network pass. LDMs work in the learned latent space, which exhibits better-scaling properties concerning the spatial dimensionality.

Therefore, latent models are efficient compared to pixel-based designs.

\subsection { Efficient Process Strategies (EPS) }

These strategies target the improvement of the diffusion process itself.  Table 2 includes some representative work in each process category that is included. A brief description of each category and its influence on the efficiency of diffusion models are discussed below:

\subsubsection{Training Strategy} To enhance sampling speed, several strategies focus on modifying the pattern of training and noise schedule. However, re-training models require more processing and increase the risk of unstable training. Fortunately, there is a family of approaches known as training-free sampling that directly augments the sample algorithm using a pre-trained model. The purpose of advanced training-free sampling is to offer an efficient sampling method for learning from a pre-trained model in fewer steps and with improved accuracy. Analytical approaches, implicit sampler, differential equation solver sampler, and dynamic programming adjustment are the three types.

By using a memory technique, dynamic programming may traverse all options to discover the optimal solution in a relatively short amount of time. In comparison to previous efficient sampling approaches, dynamic programming methods discover the optimal sample path rather than constructing strong steps that decrease error more rapidly.

\subsubsection{Noise Distribution Strategy} Unlike DDPM \cite{DDPM}, which defines noise scale as a constant, research into the effect of noise scale learning has received a lot of interest \cite{NEDM}, because noise schedule learning also counts during diffusion and sampling. Each sample step may be seen as a random walk on the direct line heading to the preceding distribution, demonstrating that noise reduction may help the sampling operation. The random walk of random noise is guided by noise learning in both the diffusion and sampling processes, resulting in more efficient reconstruction.

The underlying noise distribution of the diffusion process is Gaussian noise in most known approaches. Fitting distributions with more degrees of freedom, on the other hand, may increase the performance of such generative models. Other noise distribution forms for the diffusion process are being researched. The Denoising Diffusion Gamma Model (DDGM) \cite{DGM} demonstrates that noise from the Gamma distribution improves picture and voice creation.

The sample obtained from random noise will be tweaked anew in each sampling step to get closer to the original distribution. However, sampling with diffusion models requires too many steps, resulting in a time-consuming condition \cite{nichol2021improved}.

\subsubsection{The Mixing or Unifying Strategy} Mixed-Modeling entails incorporating another form of the generative model into the diffusion model pipeline to make use of others' high sampling speed, such as adversarial training networks and autoregressive encoders, as well as high expressiveness, such as normalizing flow \cite{Auto, DiffFlow, Diffusion-GAN}. Thus, extracting all of the strengths by combining two or more models with a specified pattern results in a possible upgrade known as Mixed-modeling.

The goal of diffusion scheme learning is to investigate the influence of different diffusion patterns on model speed. Truncating both the diffusion and sampling processes, resulting in shorter sampling time, is advantageous for lowering sampling time while enhancing producing quality. The main goal behind truncating patterns is to generate less dispersed data using various generative models such as GAN \cite{pan2019recent} and VAE \cite{razavi2019generating}.

By gradually distilling knowledge from one sample model to another, a diffusion model may be enhanced \cite{luhman2021knowledge}. Before being taught to create one-step samples as near to teacher models as possible, student models re-weight from teacher models in each distillation step. As a consequence, student models can cut the number of sample steps in half during each distillation operation.

The acceleration approach for generalized diffusion aids in the solution of a wide range of models and provides insights into effective sampling mechanisms. Other related research establishes the relationship between the diffusion model and denoising score matching, which may be considered one sort of unification.

\subsubsection{Scheduling Strategy} Improving the training schedule entails updating classic training methods such as the diffusion scheme, noise scheme, and data distribution scheme, all of which are independent of sampling.

When solving the diffusion SDE, decreasing the discretization step size helps speed up the sampling operation. Such techniques, however, would result in discretization mistakes and significantly impact model performance \cite{DiffFlow}. As a result, several strategies for optimizing the discretization scheme with a time to minimize sampling steps while maintaining excellent sample quality have been devised.

To create a prediction, the Markov process only uses the sample from the previous phase, which restricts the use of plentiful earlier data. In contrast, the transition kernel of the Non-Markovian process may rely on more samples and use more information from these samples. As a result, it can create accurate predictions with a high step size, which speeds up the sampling method.

Alternatively, by just performing certain phases of the reverse process to obtain samples, one might trade sample quality for sampling speed. Some sampling can be accomplished by pausing or truncating the forward and reverse processes early on or by retraining student networks and bypassing partial phases through knowledge distillation.

Diffusion sampling may be accomplished in a few steps with the use of strong conditioned conditions. Early Stop (ES) DDPM produced implicit distribution by producing previous data with VAE, which learned the latent space \cite{ES-DDPM}.

As previously stated, it generally takes the same number of steps for the generative process as it does for the diffusion process to reconstruct the original data distribution in DDPM \cite{DDPM}. However, the diffusion model has the so-called decoupling property in that it does not require the same number of steps for diffusing and sampling. The implicit sampling approach, which is based on the generative implicit model, includes deterministic diffusion and jump-step sampling. Surprisingly, implicit models do not require re-training since the forward's diffusion probability density is constant at all times. DDIM\cite{DDIM} uses continuous process formulation to tackle the jump-step acceleration problem.

\subsubsection{Retrieval Strategy} During training, RDMs \cite{RDM, RDM2} obtain a collection of closest neighbors from an external database, and the diffusion model is conditioned on these informative samples. Retrieval-augmentation works by looking for photos that are similar to the prompt you offer and then letting the model view them during creation.

During training, the diffusion model is fed comparable visual characteristics obtained via CLIP and from the vicinity of each training instance. By using CLIP's combined image-text embedding space \cite{shen2021much}, the model delivers very competitive performance on tasks for which it has not been explicitly trained, such as class-conditional or text-image synthesis, and may be conditioned on both text and picture embeddings improving its performance. Retrieval-Augmented Diffusion Models \cite{rombach2022text} are recently used for the text-guided synthesis of artistic images efficiently. 

Retrieval-Augmented Text-to-Image Generator (Re-Imagen), \cite{chen2022re} is a generative model that uses the extracted information to produce highly faithful images even for rare or invisible entities. At a text message, Re-Imagen accesses an external multimodal knowledge base to retrieve the relevant pairs (image, text) and uses them as references to generate the image.

\section{Comparative Performance and Discussion }
\label{sec:Compare}
In this section, we discuss  performance of different techniques, while considering the dimension of computational requirements. We also build towards the future work directions to lead new research in this exciting area. 
As noted earlier, the research focus to date has been mainly on increasing the quality of the generated samples with diffusion models.  Stable diffusion \cite{LDM} is one of the major works that changes the course of this direction by particularly  focusing on the  efficiency aspect.  Before providing the discussion, below we mention the common quality and efficiency metrics  used in the literature to benchmark diffusion model performance.

\subsection{Quality measures}
Largely agnostic to computational efficiency, these metrics fully focus on measuring the quality of the samples generated by the models. It is noteworthy that whereas these metrics are generally accepted to provide appropriate quantification of the   visual quality of images,  qualitative assessment with human visual system can still be considered the gold standard for the generated samples.

\subsubsection{Inception Score (IS)}
The inception score  was initially proposed  by  Salimans et al.~\cite{InceptionScore} to measure visual quality of generated images. To compute the  score, a pre-trained  model - commonly Inception-v3 - is used to classify the generated images. The marginal distribution of the image classes is computed by averaging the predicted probabilities for each class over all the generated images. Then,  conditional distribution of the image classes is computed.  The inception score measures the exponential of the KL divergence between the marginal and conditional distributions.
The underlying rationale behind this metric is that high quality generated images should have both high class probability and diversity.
Although the inception score can be beneficial in evaluating various generative models, it has limitations, which does not make it a widely agreed-upon evaluation metric in generative modeling~\cite{barratt2018note}.

\subsubsection{Frechet Inception Distance (FID)} Inception Score has multiple notable  limitations. Its effectiveness depends on a specific dataset, i.e., ImageNet, with 1000 classes and a pre-trained model that incorporates certain random elements such as initial weights and code structure. Consequently, there can be a bias resulting from the ImageNet model, leading to inaccurate results~\cite{kynkaanniemi2022role}. Additionally, the sample batch size used in the calculation is often much smaller than 1000 classes, which can result in unreliable statistics. Considering this, Frechet Inception Distance (FID)~\cite{FID} is also used to quantify the generated image quality.
The FID is calculated by first using a pre-trained model to extract features from both the real and generated images. The distance between the multivariate Gaussian distributions of the real and generated image features is then computed using the Frechet distance. The resulting FID score represents the difference between the real and generated image sets in terms of their visual appearance. The FID has become a popular evaluation metric for generative models, as it has been shown to correlate well with human perception of image quality.

\subsubsection{Negative Log Likelihood (NLL)}
The negative log likelihood (NLL) is a widely used loss function for training generative models, including GANs and Variational Autoencoders (VAEs). Although NLL is not truly  a standalone evaluation metric, it can be utilized as an indirect indicator to evaluate the quality of generated images. To utilize NLL for this purpose, the NLL of generated images can be calculated and compared to the same measure of the real images. The assumption is that a well-performing generative model should generate images that have negative log likelihood  comparable to those of real images.
However, relying solely on NLL for evaluating the quality of generated images is not adequate, as generative models can generate images with low NLLs but poor visual quality. In such cases, the model may have learned to memorize the training data, without generating diverse or realistic images. It is noted that we discuss NLL here for the sake of providing a more comprehensive picture. In the results discussed below, we do not include NLL scores because FID and IS are already considered more reliable than NLL scores. 
%Negative log-likelihood is viewed as a common assessment metric that describes all patterns of data distribution by Razavi et al. There has been a lot of effort on normalising flow fields [and VAE fields employ NLL as one of the assessment options. Some diffusion models, such as enhanced DDPM, consider the NLL to be the training aim.

\subsection{Efficiency measures}
As compared to the quality metrics, efficiency metrics disregard the generated image quality and fully focus on the computational aspects. We briefly review some popular efficiency measures below.
%$efficiency metrics include the following:

\subsubsection{Sampling time/steps}
Fast sampling is one the major efficiency target of the diffusion models alongside the generated image  quality. Samples generated per second is often used to quantify the model efficiency. Additionally, the measure of the number of steps taken by the model in generating these samples is also used in reporting model efficiency. Similarly, the steps taken for model training provide an estimate of training computational requirements. Normally, training and sampling steps are kept in similar ranges for optimal test time performance. Less number of steps generally imply higher efficiency.

\subsubsection{Hardware requirements} Modern High-Performance Computing (HPC) data centers are key to address the heavy computational  workload which is required for  training diffusion models. As NVIDIA  V100 and A100 GPUs are being increasingly used by the HPC clusters, many works  report the efficiency of diffusion models in terms of V100 or A100  operation days. 

\subsubsection{Model size} Often described in terms of number of model parameters, model size is also an important metric from the computational perspective. However, it is often hard to directly relate it to sample generation efficiency as less parameters do no necessarily mean fewer computations. A larger model size does have a larger memory footprint though, which indirectly affects the efficiency.

\subsection{Comparative performance}
As compared to the quality metrics, efficiency metrics are not standardized per se. We find that the existing works are often unclear about the efficiency metrics. On the other hand, standard benchmarking with the  quality metrics, especially FID scores is currently the norm. To provide an estimate of the performance of the existing methods while considering the efficiency aspect in mind, we collate the results of the most recent diffusion model techniques for the fundamental task of image synthesis.

% Table generated by Excel2LaTeX from sheet 'Sheet2'
\begin{table}[t]
\setlength{\tabcolsep}{0.1em}
  \centering
  \caption{Summary of five top performing conditional image synthesis techniques for ImageNet $256\times 256$ dataset. Diffusion models are currently providing the state-of-the-art results, outperforming other techniques, e.g., GANs and Implicit Neural Representation (INR). However, their model sizes and computational requirements are overwhelmingly large in comparison.}
    \begin{tabular}{|l|c|c|c|c|c|c|c|c|l|}
    \hline
    \multicolumn{1}{|c|}{\textbf{Method}} & \textbf{Year} & \textbf{Technique} & \textbf{FID} & \textbf{IS} & \textbf{Steps/time} & \textbf{Model size} & \textbf{Hardware} & \textbf{Train time} & \multicolumn{1}{c|}{\textbf{Comment}} \\
    \hline
    MDT-XL/2~\cite{MDT} & 2023  & Diffusion & \textcolor[rgb]{ .129,  .145,  .161}{1.79} & \textcolor[rgb]{ .129,  .145,  .161}{283.01} &   -    & 675.8 M & 8 x Nvidia A100  & 33 hr for MDT-S &  3x faster than MDT/DiT~\cite{MDTDIT} \\
    \hline
    ViT-XL~\cite{ViTXL} & 2023  & Diffusion & 2.06  &    -   & 2.1M  & 451M  &   -    &   -    & $3.4\times$ faster convergence than baseline \\
    \hline
    DiT-XL/2~\cite{MDTDIT} & 2022  & Diffusion & 2.27  & 278.24 & 7M    & 675M  & JAX TPU v3-256 pods & 5.7 sec/itr & Uses diffusion transformers \\
    \hline
    S-GAN-XL~\cite{StyleGANXL} & 2022  & GAN   & 2.3   &       & 0.07 sec & 166.3M & 4 x Nvidia V100  &   -    & High speed inference \\
    \hline
    Poly-INR~\cite{PolyINR} & 2023  & INR & \textcolor[rgb]{ .129,  .145,  .161}{2.86} &   -    & 0.05 sec & 46M   & Nvidia-RTX-6000 &   -    & Much smaller model and fast inference \\
    \hline
    \end{tabular}%
  \label{tab:ImageNetSynthesis}%
\end{table}%

% Table generated by Excel2LaTeX from sheet 'Sheet2'
\begin{table}[t]
  \centering
  \setlength{\tabcolsep}{0.1em}
  \caption{Summary of five top performing conditional image synthesis techniques for CelebA $64\times 64$ dataset.}
    \begin{tabular}{|l|c|c|c|c|c|c|c|l|}
    \hline
    \multicolumn{1}{|c|}{\textbf{Method}} & \textbf{Year} & \textbf{Technique} & \textbf{FID} & \textbf{Steps/time} & \textbf{Model size} & \textbf{Hardware} & \textbf{Train time} & \multicolumn{1}{c|}{\textbf{Comment}} \\
    \hline
    DDPM-IP~\cite{DDPMIP} & 2023  & Diffusion & 1.31  & 1000  & 295M  & 16 x Nvidia V100 & 5 days & Enhances DDPM with input perturbation  \\
    \hline
    STDDPM-G++~\cite{EDMG} & 2022  & Diffusion & 1.34  & 1000  &    -   & Nvidia V100 &   -    & Fine-tunes  pretrained models \\
    \hline
    Diffusion StyleGAN2~\cite{Diffusion-GAN} & 2022  & Diff.+GAN & 1.69  &  -     &   -    & 4 or 8 Nvidia V100  &  -     & Has diffusion discriminator and generator \\
    \hline
    INDM~\cite{INDM}  & 2022  & Diffusion & 1.75  &   -    &    -   & P40(96Gb) 4 or 8 & 4+5 days & Combines normalizing flow and diffusion \\
    \hline
    Soft Diffusion~\cite{Soft} & 2022  & Diffusion & 1.85  &    -   &   -    & 16 x v2-TPUs & 6 iterations/sec & Assumes diffusion operator is known \\
    \hline
    \end{tabular}%
  \label{tab:Celeb}%
\end{table}%
\begin{table}[t]
  \centering
  \caption{Summary of five top performing  image synthesis techniques for CIFAR-10. }
    \begin{tabular}{|l|c|c|c|c|c|l|}
    \hline
    \multicolumn{1}{|c|}{\textbf{Method}} & \textbf{Year} & \textbf{Technique} & \textbf{FID} & \textbf{Model size} & \textbf{Hardware} & \multicolumn{1}{c|}{\textbf{Comment}} \\
    \hline
    EDM-G++~\cite{EDMG} & 2022  & Diffusion & \textcolor[rgb]{ .129,  .145,  .161}{1.64} &       & Nvidia V100 & Conditional generation \\
    \hline
    PFGM++~\cite{PFGM} & 2023  & Diffusion & \textcolor[rgb]{ .129,  .145,  .161}{1.74} &    -   & 4 or 8 Nvidia V100  & Unifies Poisson Flow Gen. Models with Diffusion \\
    \hline
    EDM-G++~\cite{EDMG} & 2022  & Diffusion & \textcolor[rgb]{ .129,  .145,  .161}{1.77} &    -   & Nvidia V100 & Unconditional generation \\
    \hline
    StyleGAN-XL~\cite{StyleGANXL} & 2022  & GAN   & 1.85  & 166.3M & 4 x Nvidia V100 & Very high speed generation because of GAN \\
    \hline
    STF~\cite{STF}   & 2023  & Diffusion & \textcolor[rgb]{ .129,  .145,  .161}{1.90} &     -  & 2 x Nvidia A100 & Unconditional generation \\
    \hline
    \end{tabular}%
  \label{tab:CIFAR}%
\end{table}%

In Tables~\ref{tab:ImageNetSynthesis}, \ref{tab:Celeb} and \ref{tab:CIFAR}, we respectively summarize the details of the top performing state-of-the-art image synthesis/generation techniques for the popular benchmarks of ImageNet $256\times 256$~\cite{ImageNet}, Celeb-A~\cite{CelebA} and CIFAR-10~\cite{CIFAR10}. Synthesizing images is the most fundamental task for generative visual models. Hence, it provides a clear picture of the generative abilities of the underlying techniques. In the tables, we report the top performing methods based on the FID score. We do not restrict the methods to pertain to diffusion models only. It can be seen that for all the datasets, diffusion model based techniques have now established the state-of-the-art results. It is becoming increasingly rare to encounter GAN-based methods to outperform diffusion-based techniques in terms of quantitative metrics.  

The critical aspect for the diffusion-based techniques relates to their computational requirements. It can be seen that the diffusion techniques not only demand high-end resources, but also many days of training (in general) on such resources. Moreover, the model sizes of diffusion methods are often quite large. Comparatively, GAN-based methods and other techniques, e.g., Implicit Neural Modeling (INM)~\cite{PolyINR} have much smaller models and resource requirements - see Table~\ref{tab:ImageNetSynthesis}. Due to the high-end resource demand, diffusion modeling research is still  more restricted to the research labs that have access to HPC facilities. Despite a large interest of the research community in diffusion models, this bottleneck still hinders the true democratization of diffusion models. A potential direction to overcome this hurdle can be in the form of organizing public challenges to benchmark efficiency of the diffusion models. Or encourage resource-limited image synthesis.  This is currently an open gap, which is likely to attract the attention of the community once the results on quality metrics start to saturate.

%This is another direction to contribute to diffusion model efficiency research. 
Similar to the fundamental task of image generation, image inpainting is also a natural application of generative modeling, among many others. It has recently attracted significant attention, partially thanks to  the rise of generative image synthesis models \cite{glide, ILVR, NTIRE, palette}. Most inpainting solutions perform well on object removal or texture synthesis, while semantic generation is still difficult to achieve. To facilitate image inpainting research, NTIRE 2022 \cite {NTIRE} Image Inpainting Challenge was introduced with the target to develop solutions that can achieve a robust performance across different and challenging masks while generating compelling semantic images. As noted in the challenge report~\cite{NTIRE}, the winner of track-2 of the challenge also relies on the Latent Diffusion Model~\cite{LDM}. It can be easily anticipated that in the near future, diffusion models will be able to completely replace techniques such as GANs in image synthesis related tasks, provided that their computational demands remains in check.

%%%%%%%%%%%%%%%%%%%%%%%%%%%%
% \begin{table}[]
% \begin{center}
% \begin{tabular}{|l|l|l|l|l|}
% \hline
% Model     & Year & FID  & Steps & Parameters M \\ \hline
% PGGAN     & 2017 & 8.03 & NA    & 23.1         \\ \hline
% DDGAN     & 2021 & 7.64 & NA    &   -           \\ \hline
% StyleSwin & 2021 & 3.25 & NA    &   -           \\ \hline
% ADM       & 2021 & 1.9  & 1000  &   -           \\ \hline
% LDM       & 2022 & 2.95 & 200   & 1.9  \\ \hline       
% \end{tabular}

%     \end{center}
% \end{table}

% \begin{table}[]
% \begin{center}
% \begin{tabular}{|l|l|l|l|}
% \hline
% Model       & FLOP   & Parameters & Inference Time \\\hline
% LDM-8       & 37.1 G & 589.8 M    & 0.82547         \\\hline
% Frido       & 37.3 G & 1.179 B    & 1.02865         \\\hline
% Dido-gating & 39.7 G & 697.8 M    & 0.91782    \\ \hline  
% \end{tabular}
% \captionsetup{singlelinecheck=false, font=small, labelfont=bf}
%   \caption{Efficiency comparison of best performing works on challenging Image Generation datasets ImageNet (Top) and  COCO (Bottom) in terms of reported efficiency metrics for diffusion models in the literature. }
%     \end{center}
% \end{table}
%%%%%%%%%%%%%%%%%%%%%

\subsection{Further discussion}
\begin{figure*}[t]
\centering
\includegraphics[width=0.7\textwidth]{ 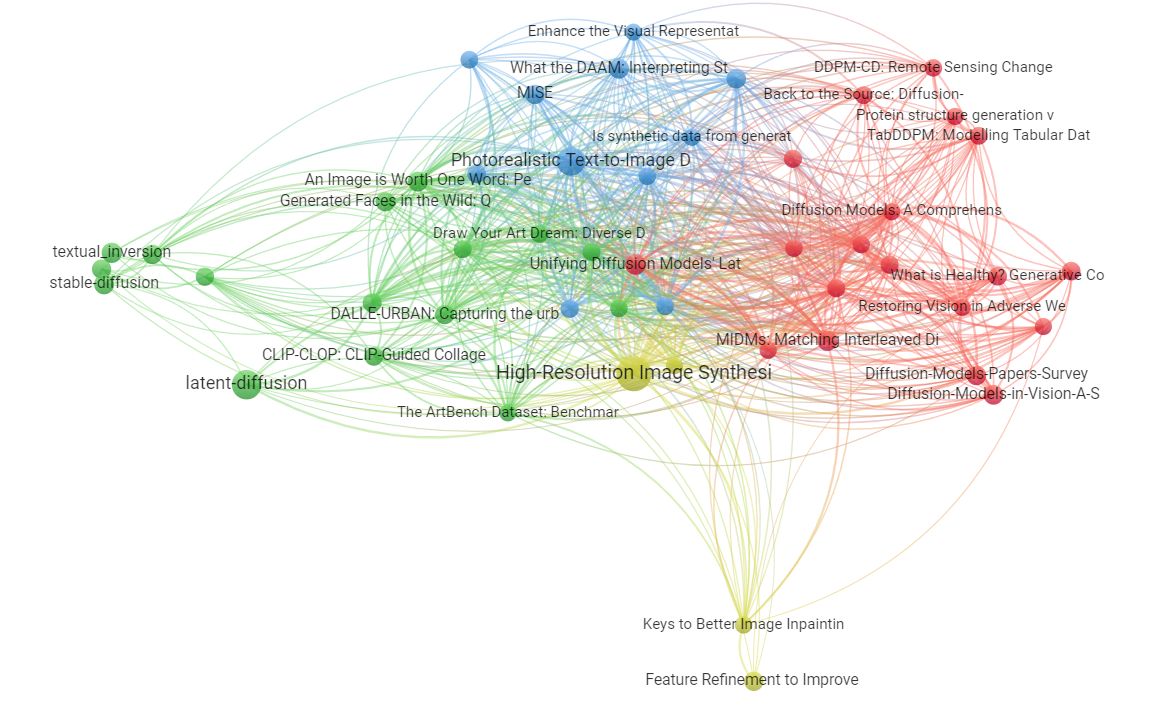} 
\caption{Visualization of Bibliometric Networks for assessing the impact of latent diffusion models" It shows the top 50 papers in terms of their relevance and impact on the topic and with each other. Each cluster provides a distinct theme and is represented by the same color dots. Mutual connections in the network are based on their term similarities. }
\label{fig:Bib2}
\end{figure*}

As noted earlier, Latent Diffusion Model~\cite{LDM} is one of the pioneering works that conscientiously accounts for the efficiency aspect.    To find the impact of LDM on the emerging trends in the literature, we use a bibliographic network. This network provides an  insight into the impact of LDM on the emerging literature. This, in turn identifies the response of the research community to account for the computational aspects of diffusion models.
For the bibliographic network, we use a clustering approach. In the cluster analysis, the number of sub-problems is set by the resolution. The greater the value of this parameter, the more clusters will be created. We use a small number of clusters to focus on the most representative works in terms of relevance and impact, which resulted in four major clusters based on 50 representative papers. Figure~\ref{fig:Bib2} displays these clusters, and Table~\ref{tab:cluster} lists representative papers in each cluster for reference.

%Such visualization of a bibliometric network provides an automated insight into relevant literature. % that can not be figured out manually. This visualization and its depth of understanding helped us to revise our taxonomies, which are discussed in the following sections. 
\begin{table}[t]
    \centering
    \begin{tabular}{|l|l|}
     \hline
       \textbf{Cluster}  &  \textbf{Representative paper title} \\ \hline
       \textcolor{yellow}{Yellow}  & High-Resolution Image Synthesis with Latent Diffusion Models \\ \hline
       \textcolor{green}{Green} & Text-Guided Synthesis of Artistic Images with Retrieval-Augmented Diffusion Models \\ \hline
       \textcolor{blue}{Blue} &  What the DAAM: Interpreting Stable Diffusion Using Cross Attention \\ \hline
       \textcolor{red}{Red} & Unifying Diffusion Models’ Latent Space, with Applications to Cycle Diffusion and Guidance \\ \hline
    \end{tabular}
    \caption{Representative paper titles from the bibiliometric network in Fig.~\ref{fig:Bib2}. }
    \label{tab:cluster}
\end{table}

\section{Emerging and future directions}
\label{sec:Compare}
Diffusion models in computer vision are gaining large attention due to their popularity, usability, and creativity. This is particularly evident after the efficient utilization of computing resources by LDM~\cite{LDM} and the availability of stable diffusion in open source. The impact of stable diffusion can be considered transformative. Nonetheless, the scientific literature is also witnessing new research endeavors directed towards other challenges on a regular basis. Emerging and prospective research directions in this domain are briefly outlined below.
%The popularity, usability, and creativity of diffusion models are attracting  new research efforts in  computer vision, especially after the efficient use of computing resources and open source availability of stable diffusion. It is fair to say that stable diffusion has proven to be a game-changing technique. However, novel  contributions are emerging every day in the literature that are also directed to other challenges. Some of the emerging and future research directions in this area are summarized below.
\begin{itemize}

\item Adversarial manipulation of deep learning models is currently a highly active research direction~\cite{Akhtar}. With the impressive generative abilities of diffusion models, it is likely that these model will attract a significant attention of the research community from the adversarial perspective. We can anticipate that this manipulation will not only be limited to cause these models to generate unexpected outputs, but also reveal details of the training data. Considering the downstream applications of diffusion models in public domains, training data privacy preservation is expected to be tested thoroughly by the attackers with adversarial manipulation.   

\item Due to impressive generative abilities of the diffusion model, it can be expected that this technology will be exploited for DeepFakes. Hence, identification of generated vs real images in the context of diffusion models is also a potential venue that is likely to be explored thoroughly in the future.

\item It is known that adversarial inputs can be cleaned via denoising process~\cite{AkhtarCVPR}, since diffusion models are inherently based on denoising, it can be anticipated that their role in developing adversarial defenses can become significant in the future.

\item  From the viewpoint of efficiency, it can be anticipated that interpretability and explainability of diffusion models can come into play. If the actual underlying  model details are  well-interpreted,  it can lead to efficient diffusion model designs.  An interpretability method called DAAM      \cite{wu2022unifying} is already introduced to produce pixel-level attribution maps based on upscaling and aggregating cross-attention activations in the latent denoising subnetwork.

\item Another emerging direction is the development of few-shot diffusion models, which present a framework for few-shot generation leveraging conditional DDPMs. These models can be trained to adapt the generative process conditioned on a small set of images from a given class. An approach like DreamBooth \cite{DreamBooth} can provide  "personalization" of text-to-image diffusion models (specializing them to users' needs) on a similar concept. Given as input just a few images of a subject, such models can fine-tune a pre-trained text-to-image model such that it learns to bind a unique identifier with that specific subject.  By leveraging the semantic prior embedded in the model with a class-specific prior preservation loss, these models enable synthesizing the subject in diverse scenes, poses, views, and lighting conditions that do not appear in the reference images. CycleDiffusion is introduced by \cite{tang2022daam} that shows that large-scale text-to-image diffusion models can be used as zero-shot image-to-image editors. It can guide pre-trained diffusion models and GANs by controlling the latent codes in a unified plug-and-play formulation, based on energy-based models.

%\item Retrieval-augmentation works by looking for images similar to the specified prompt and then the model can see them during generation.

\item In the past, most text-to-image models were developed as propriety applications. However, the coming of stable diffusion open source has initiated another trend that will help evolve diffusion research.

\item  Another emerging research direction is innovative architectures for video diffusion models \cite{ho2022video,yang2022diffusion} which is a natural extension of the image architectures. These architectures may use joint training from image and video data to generate long and higher-resolution videos. Generation of video without text-video data can introduce efficient designs \cite{singer2022make}.

\item Diffusion models are also promising candidates in many other applications. Hence, we can expect their appearance in a wide range of applications. For instance,  Motion Diffusion Model \cite{action1} is a carefully tuned classifier-free generative diffusion-based model  introduced for the human motion domain. The model is based on a transformer and combines knowledge from the motion generation literature. 
%It uses sample prediction rather than noise at each scattering stage. This facilitates the use of established geometric losses at motion locations and velocities, such as loss of foot contact. 
It is a generic approach that allows different conditioning modes and different generation tasks. Similar work is Motiondiffuse \cite{action2} which is text-driven human motion generation with a diffusion model. Such methods indicate a future trend to generate complex nature of visual data with diffusion models.

\end{itemize}

%%%%%%%%%%%%%%%%%%%%%%%%%%%%%%%%%%%%%%%%%%%%%%%%%%%%%%%%%%%%
\section{Conclusion}
\label{sec:Conc}
%In this review, we presented the most recent advances in diffusion models and discussed important design aspects that directly impact the computational efficiency of the diffusion models. We focused on recently proposed design choices that have resulted in efficient diffusion models. Unlike previous works that categorize diffusion models generally, this article particularly focuses on the aspects related to computational requirements. Generally, current diffusion models are mainly relying on high-end computational resources. We anticipate that the future models will particularly address the computational efficiency. Developments along this direction can further catapult the widespread of diffusion models due to their impressive generative abilities. We have also provided a brief comparative analysis of the top performing existing  approaches in image synthesis, which clearly shows that diffusion models have already replaced GANs as the top performers. However, this success is currently relying strongly on high computational requirements.

This review article presents recent advances in diffusion models and discusses key design aspects that have a direct impact on their computational efficiency. The article specifically focuses on design choices that have led to efficient diffusion models, in contrast to previous works that generally categorized diffusion models. Currently, diffusion models heavily rely on high-end computational resources, but it is expected that future models will address computational efficiency to enable wider use of diffusion models due to their impressive generative abilities. The article includes a brief comparative analysis of the top performing existing approaches in image synthesis, demonstrating that diffusion models have surpassed GANs as the top performers. However, the success of diffusion models is currently heavily dependent on high computational requirements.

%in terms of efficiency metrics and provided new directions for future research work regarding computationally efficient diffusion models.



%\begin{thebibliography}{1}


\bibliographystyle{IEEEtran}
%\bibliography{IEEEabrv}
%\end{thebibliography}

%\begin{biography}[{\includegraphics[width=1in,height=1.25in,clip,keepaspectratio]{mshell}}]{Michael Shell}

% ==== SWITCH OFF the BIO for submission
% ==== SWITCH OFF the BIO for submission
\begin{IEEEbiography}[{\includegraphics[width=1in,height=1.25in,clip,keepaspectratio]{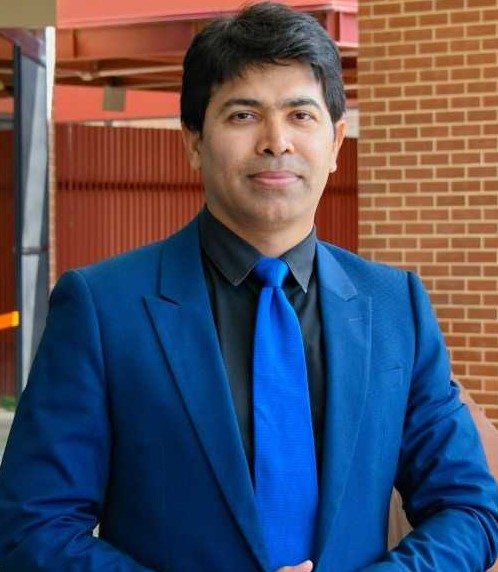}}]{Anwaar Ulhaq}
 holds a PhD (Artificial Intelligence) from Monash University, Australia. He is working as a senior lecturer (AI) at the school of computing, Mathematics, and Engineering at Charles Sturt University, Australia. He has developed national and international recognition in computer vision and image processing. His research has been featured 16 times in national and international news venues, including ABC News and IFIP (UNESCO). He is an active member of IEEE, ACS and the Australian Academy of Sciences. As Deputy Leader of the Machine Vision and Digital Health Research Group (MaViDH), he provides leadership in Artificial Intelligence research and leverages his leadership vision and strategy to promote AI research by mentoring junior researchers in AI and supervising HDR students devising plans to increase research impact.
\end{IEEEbiography}
\begin{IEEEbiography}[{\includegraphics[width=1in,height=1.25in,clip,keepaspectratio]{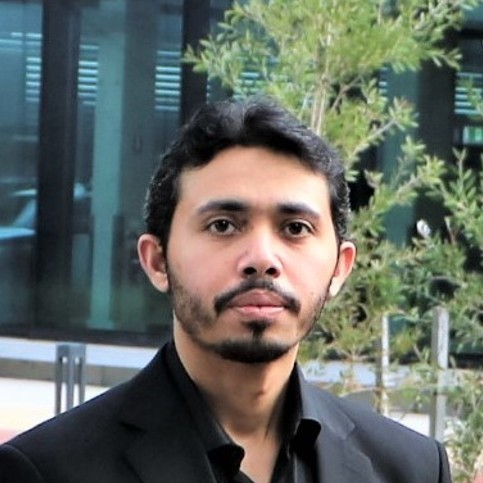}}]{Naveed Akhtar}
 is a Senior Research Fellow at the University of Western Australia (UWA). He received his PhD in Computer Science from UWA and Master degree  from Hochschule Bonn-Rhein-Sieg, Germany. He is a recipient of the prestigious fellowship by the Australian Office of National Intelligence. He is also a finalist of the Western Australia's Early Career Scientist of the Year 2021 and Universal Scientific Education and Research Network top 10 young scientist in Formal Sciences for 2021.  He is an ACM Distinguished Speaker and serves an Associate Editor of IEEE Transactions on Neural Network and Learning Systems. He has also served as an Associate Editor for IEEE Access and Guest Editor of Neural Computing and Applications and Remote Sensing journals. 
His research is regularly published in the prestigious sources of computer vision, including IEEE Trans.~Pattern Analysis and Machine Intelligence, IEEE Conf.~on Computer Vision and Pattern Recognition (CVPR) and European Conference on Computer Vision (ECCV). He also served as an Area Chair for CVPR'22 and ECCV'22.  
%During his PhD, he was  recipient of multiple scholarships, and  runner-up for the Canon Extreme Imaging Competition in 2015. 
\end{IEEEbiography}

%% insert where needed to balance the two columns on the last page with
%%\newpage

% ==== SWITCH OFF the BIO for submission
% ==== SWITCH OFF the BIO for submission


\vfill


\end{document}